\documentclass[runningheads]{llncs}
\usepackage{graphicx}
\usepackage{tikz}
\usepackage{comment}
\usepackage{amsmath,amssymb} 
\usepackage{color}
\usepackage{xcolor}
\usepackage{colortbl}
\usepackage{enumitem}
\usepackage{bm}
\usepackage{hyperref}

\usepackage[accsupp]{axessibility}
\usepackage[normalem]{ulem}

\usepackage{booktabs}

\usepackage{caption}
\usepackage{subcaption}

\newcommand\myparagraph[1]{\paragraph{\textbf{#1}}}

\begin{document}
\pagestyle{headings}
\mainmatter
\def\ECCVSubNumber{83}  

\title{CAR: Class-aware Regularizations for \\Semantic Segmentation } 

\titlerunning{CAR: Class-aware Regularizations for Semantic Segmentation}
%
\author{Ye Huang\inst{1} \and
Di Kang\inst{2} \and
Liang Chen\inst{3} \and
Xuefei Zhe\inst{2} \and \\
Wenjing Jia\inst{1} \and
Linchao Bao\inst{2} \and
Xiangjian He\inst{4}\thanks{Corresponding author}
}
\authorrunning{Y. Huang et al.}
%
\institute{University of Technology Sydney, Australia \and
Tencent AI Lab \and
Fujian Normal University \and
University of Nottingham Ningbo China\\}
\maketitle

\begin{abstract}
Recent segmentation methods, such as OCR and CPNet, utilizing ``class level'' information in addition to pixel features, have achieved notable success for boosting the accuracy of existing network modules. 
However, the extracted class-level information was simply concatenated to pixel features, without explicitly being exploited for better pixel representation learning.
Moreover, these approaches learn soft class centers based on coarse mask prediction, which is prone to error accumulation. 
In this paper, aiming to use class level information more effectively, 
we propose a universal Class-Aware Regularization (CAR) approach to optimize the intra-class variance and inter-class distance during feature learning, motivated by the fact that humans can recognize an object by itself no matter which other objects it appears with.
Three novel loss functions are proposed. 
The first loss function encourages more compact class representations within each class,
the second directly maximizes the distance between different class centers,
and the third further pushes the distance between inter-class centers and pixels.
Furthermore, the class center in our approach is directly generated from ground truth instead of from the error-prone coarse prediction.
Our method can be easily applied to most existing segmentation models during training, including OCR and CPNet, and can largely improve their accuracy at no additional inference overhead. 
Extensive experiments and ablation studies conducted on multiple benchmark datasets demonstrate that the proposed CAR can boost the accuracy of all baseline models by up to 2.23\% mIOU with superior generalization ability.
The complete code is available at  \href{https://github.com/edwardyehuang/CAR}{https://github.com/edwardyehuang/CAR}.

\keywords{Class-aware regularizations, semantic segmentation}

\end{abstract}

\section{Introduction}

Semantic segmentation, which assigns a class label for each pixel in an image, is a fundamental task in computer vision. 
It has been widely used in many real-world scenarios that require the results of scene parsing for further processing, \textit{e.g.}, image editing, autopilot, etc. 
It also benefits many other computer vision tasks such as object detection and depth estimation.

After the early work FCN~\cite{cFCN} which used fully convolutional networks to make the dense per-pixel segmentation task more efficient,
many works~\cite{cPSPNet,cDeepLab} have been proposed which have greatly advanced the segmentation accuracy on various benchmark datasets.
Among these methods, many of them have focused on better fusing spatial domain context information to obtain more powerful feature representations (termed \emph{pixel features} in this work) for the final per-pixel classification.
For example, VGG~\cite{cVGG} utilized large square context information by successfully training a very deep network, and
DeepLab~\cite{cDeepLab} and PSPNet~\cite{cPSPNet} utilized multi-scale features with the ASPP and PPM modules.

Recently, methods based on dot-product self-attention (SA) have become very popular since they can easily capture the long-range relationship between pixels ~\cite{cNonLocal,cDualAttention,cOCNet,cCFNet,cEMANet,cANNN,cViT,cDPT,cSETR}. 
SA aggregates information dynamically (by different attention maps for different inputs) and selectively (using weighted averaging spatial features according to their similarity scores). 
Using multi-scale and self-attention techniques during spatial information aggregation has worked very well (\textit{e.g.}, 80\% mIOU on Cityscapes~\cite{cCityScapes} (single-scale w/o flipping)).

As complements to the above methods, many recent works have proposed various modules to utilize class-level contextual information. 
The class-level information is often represented by the class center/context prior which are the mean features of each class in the images.
OCR~\cite{cOCR} and ACFNet~\cite{cACFNet} extract ``soft'' class centers according to the predicted coarse segmentation mask by using the weighted sum.
CPNet~\cite{cCPN} proposed a context prior map/affinity map, which indicates if two spatial locations belong to the same class, and used this predicted context prior map for feature aggregation.
However, they~\cite{cOCR,cACFNet,cCPN} simply concatenated these class-level features with the original pixel features for the final classification.

In this paper, we also focus on utilizing class level information. 
Instead of focusing on how to better extract class-level features like the existing methods~\cite{cOCR,cACFNet,cCPN},
we use the simple, but accurate, average feature according to the GT mask, and focus on maximizing the inter-class distance during feature learning. This is because it mirrors how humans can robustly recognize an object by itself no matter what other objects it appears with.

\begin{figure}[h]
\centering
\includegraphics[width=1.0\linewidth]{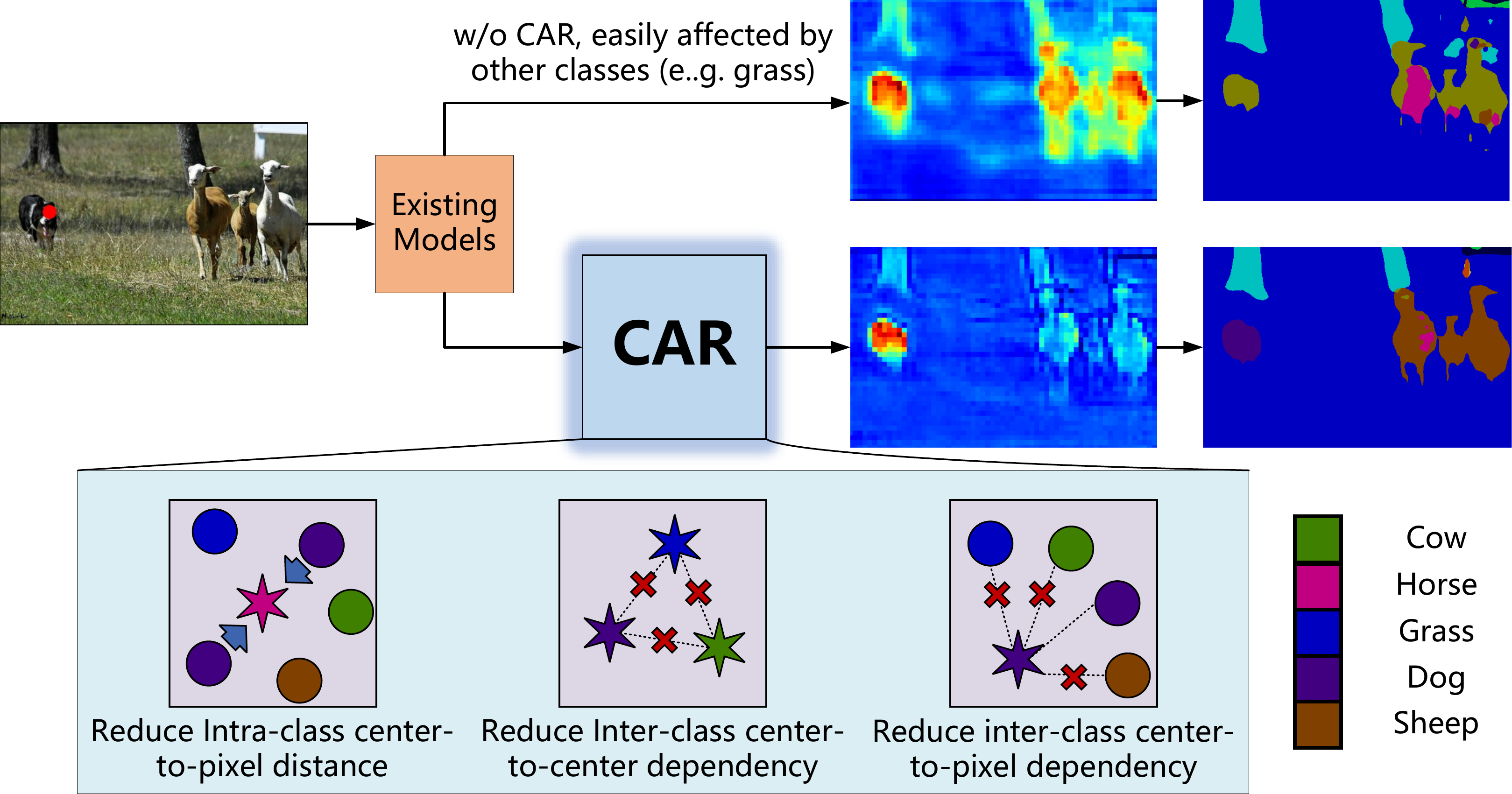}
\caption{The concept of the proposed CAR. 
Our CAR optimizes existing models with three regularization targets: 
1) reducing pixels' intra-class distance, 
2) reducing inter-class center-to-center dependency, and 
3) reducing pixels' inter-class dependency. 
As highlighted in this example (indicated with a red dot in the image), with our CAR, the grass class does not affect the classification of dog/sheep as much as before, and hence
successfully avoids previous (w/o CAR) mis-classification.
}
\label{fig:CAR:Intro}
\end{figure}

Learning more separable features makes the features of a class less dependent upon other classes, resulting in improved generalization ability, especially when the training set contains only limited and biased class combinations (\textit{e.g.}, cows and grass, boats and beach). 
Fig.~\ref{fig:CAR:Intro} illustrates an example of such a problem,
where the classification of dog and sheep depends on the classification of grass class, and has been mis-classified as cow.  
In comparison, networks trained with our proposed CAR successfully generalize to these unseen class combinations.

To better achieve this goal, 
we propose CAR, a class-aware regularizations module, that optimizes the class center (intra-class) and inter-class dependencies during feature learning. 
Three loss functions are devised: the first encourages more compact class representations within each class, and the other two directly maximize the distance between different classes. 
Specifically, an intra-class center-to-pixel loss (termed as ``intra-c2p'', Eq.~\eqref{eq:CAR:intra_diff}) is first devised to produce more compact representation within a class by minimizing the distance between all pixels and their class center. 
In our work, a class center is calculated as the averaged feature of all pixels belonging to the same class according to the GT mask. 
More compact intra-class representations leave a relatively large margin between classes, thus contributing to more separable representations. 
Then, an inter-class center-to-center loss (``inter-c2c'', Eq.~\eqref{eq:CAR:inter-c2c}) is devised to maximize the distance between any two different class centers.
This inter-class center-to-center loss alone does not necessarily produce separable representations for every individual pixels. 
Therefore, a third inter-class center-to-pixel loss (``inter-c2p'', Eq.~\eqref{eq:CAR:inter_c2p}) is proposed to enlarge the distance between every class center and all pixels that do not belong to the class. 

In summary, our contributions in this work are:
\begin{enumerate}[topsep=0pt,itemsep=0pt,parsep=0pt,partopsep=0pt]
\item We propose a universal class-aware regularization module that can be integrated into various segmentation models to largely improve the accuracy. 
%
\item We devise three novel regularization terms to achieve more separable and less class-dependent feature representations 
by minimizing the intra-class variance and maximizing the inter-class distance.
\item We calculate the class centers directly from ground truth during training, thus avoiding the error accumulation issue of the existing methods and introducing no computational overhead during inference.
\item We provide image-level feature-similarity heatmaps to visualize the learned inter-class features with our CAR are indeed less related to each other. 
\end{enumerate}

\section{Related Work}

\textbf{Self-Attention.}
Dot-product self-attention proposed in~\cite{cNonLocal,cAttentionIsAllYourNeed}  has been widely used in semantic segmentation~\cite{cDualAttention,cOCNet,cCFNet,cANNN}. 
Specifically, self-attention determines the similarity between a pixel with every other pixel in the feature map by calculating their dot product, followed by softmax normalization. 
With this attention map, 
the feature representation of a given pixel is enhanced by aggregating features from the whole feature map weighted by the aforementioned attention value, thus easily taking long-range relationship into consideration and yielding boosted performance. 
In self-attention, in order to achieve correct pixel classification, the representation of pixels belonging to the same class should be similar to gain greater weights in the final representation augmentation.

\noindent\textbf{Class Center.}
In 2019~\cite{cACFNet,cOCR}, the concept of \emph{class center} was introduced to describe the overall representation
of each class from the categorical context perspective. 
In these approaches, the center representation of each class was determined by calculating the dot product of the feature map and the coarse prediction (\textit{i.e.}, weighted average) from an auxiliary task branch, supervised by the ground truth~\cite{cPSPNet}. 
After that, those intra-class centers are assigned to the corresponding pixels on feature map.
Furthermore, in 2020~\cite{cCPN}, a learnable kernel and one-hot ground truth were used to separate the intra-class center from inter-class center, and then concatenated with the original feature representation.

All of these works~\cite{cOCR,cACFNet,cCPN} have focused on extracting the intra (inter) class centers, but they then simply concatenated the resultant class centers with the original pixel representations to perform the final logits. 
We argue that the categorical context information can be utilized in a more effective way so as to reduce the inter-class dependency. 

\begin{figure}[t]
\centering
\begin{subfigure}[b]{0.48\textwidth}
    \centering
    \includegraphics[width=\textwidth]{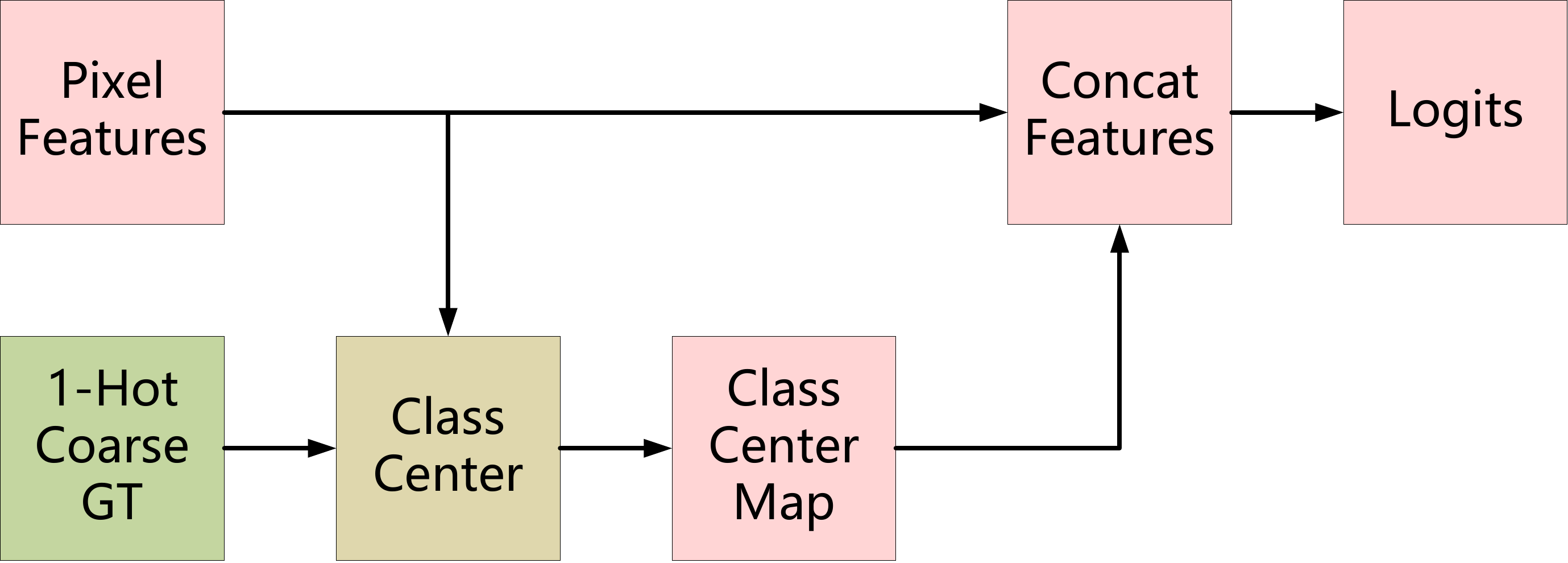}
    \caption{Design of OCR, ACFNet and CPNet}
\end{subfigure}
\hfill
\begin{subfigure}[b]{0.48\textwidth}
    \centering
    \includegraphics[width=\textwidth]{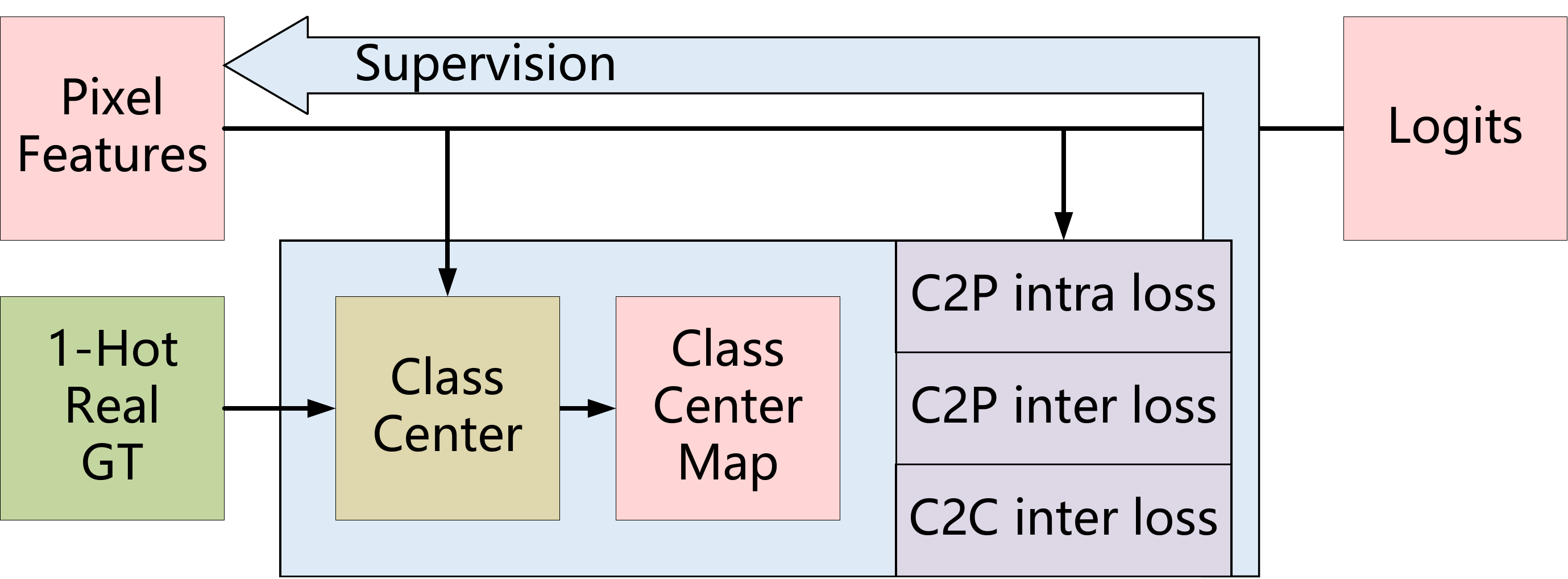}
    \caption{Our CAR}
\end{subfigure}
\hfill
\caption{The difference between the proposed CAR and previous methods that use class-level information.
Previous models focus on extracting class center while using simple concatenation of the original pixel feature and the class/context feature for later classification.
In contrast, our CAR uses direct supervision related to class center as regularization during training,
resulting in small intra-class variance and low inter-class dependency. 
See Fig.~\ref{fig:CAR:Intro} and Sec.~\ref{CAR:sec:method} for details.}
\label{fig:CAR:CompareOCR}
\end{figure}

To this end, we propose a CAR approach, where the extracted class center is used to directly regularize the feature extraction process so as to boost the differentiability of the learned feature representations (see Fig.~\ref{fig:CAR:Intro}) and reduce their dependency on other classes. 
Fig.~\ref{fig:CAR:CompareOCR} contrasts the two different designs. 
More details of the proposed CAR are provided in Sect.~\ref{CAR:sec:method}.

\noindent\textbf{Inter-Class Reasoning.}
Recently,~\cite{cHANet,cDependencyNet} studied the class dependency as a dataset prior and demonstrated that inter-class reasoning could improve the classification performance. 
For example, a car usually does not appear in the sky, 
and therefore the classification of sky can help reduce the chance of mis-classifying an object in the sky as a car.
However, due to the limited training data, such class-dependency prior may also contain bias, especially when the desired class relation rarely appears in the training set. 

Fig.~\ref{fig:CAR:Intro} shows such an example. 
In the training set, cow and grass are dependent on each other. However, as shown in this example, when there is a dog or sheep standing on the grass, the class dependency learned from the limited training data may result in errors and predict the target into a class that appears more often in the training data, \textit{i.e.}, cow in this case. 
In our CAR, we design inter-class and intra-class loss functions to reduce such inter-class dependency and achieve more robust segmentation results.

\section{Methodology}
\label{CAR:sec:method}
\begin{figure}[!htbp]
\centering
\includegraphics[width=0.96\linewidth]{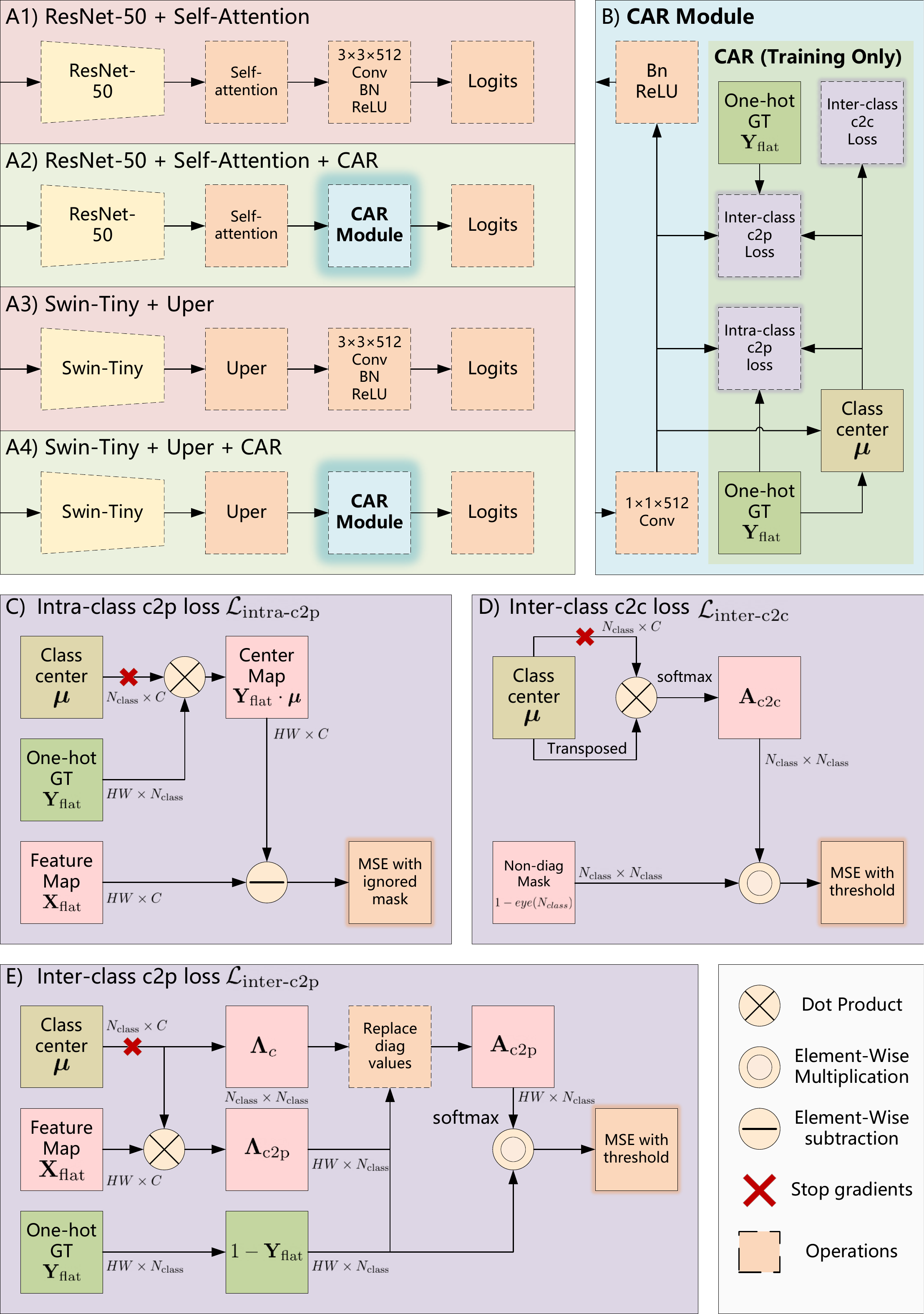}
\caption{\textbf{The proposed CAR approach.} 
CAR can be inserted into various segmentation models, right before the logit prediction module (A1-A4). 
CAR contains three regularization terms, including 
(C) intra-class center-to-center loss $\mathcal{L}_{\text{intra-c2p}}$ (Sec.~\ref{sec:intra-c2p}), 
(D) inter-class center-to-center loss $\mathcal{L}_{\text{inter-c2c}}$ (Sec.~\ref{sec:inter_c2c}), 
and (E) inter-class center-to-pixel loss $\mathcal{L}_{\text{inter-c2p}}$ (Sec.~\ref{sec:inter_c2p}). 
}
\label{fig:CAR:Arch}
\end{figure}

\subsection{Extracting Class Centers from Ground Truth}

Denote a feature map and its corresponding resized one-hot encoded ground-truth mask as 
$\mathbf{X} \in \mathbb{R}^ {H\times W\times C}$\footnote{$H$, $W$ and $C$ denote images' height and width, and number of channels, respectively.} 
and 
$\mathbf{Y} \in  \mathbb{R}^ {H\times W\times N_\text{class}}$, respectively.
%
%
We first get the spatially flattened class mask~$\mathbf{Y}_{\text{flat}} \in  \mathbb{R}^ {HW \times N_\text{class}}$ and flattened feature map~$\mathbf{X}_{\text{flat}} \in  \mathbb{R}^ {HW \times C}$. 
Then, the class center\footnote{It is termed as \emph{class center} in~\cite{cACFNet} and \emph{object region representations} in~\cite{cOCR}.}, 
which is the average features of all pixel features of a class, can be calculated by:
\begin{equation}
    \boldsymbol{\mu}_{image} = \frac{\mathbf{Y}_{\text{flat}}^{T}\cdot\mathbf{X}_\text{flat}}{\mathbf{N}_{\text{non-zero}}} \in \mathbb{R}^{N_\text{class} \times C},
\end{equation}
where $ \mathbf{N}_{\text{non-zero}} $ denotes the number of non-zero values in the corresponding map of the ground-truth mask $\mathbf{Y}$.
In our experiments, to alleviate the negative impact of noisy images, we calculate the class centers using all the training images in a batch, and denote them as $\boldsymbol{\mu_\text{batch}}$\footnote{We use $\boldsymbol{\mu}$ and omit the subscript $batch$ for clarity.}.

\subsection{Reducing Intra-Class Feature Variance}

\subsubsection{Motivation.}
More compact intra-class representation can lead to a relatively larger margin between classes, and therefore result in more separable features.
In order to reduce the intra-class feature variance, existing works~\cite{cNonLocal,cDualAttention,cANNN,cCPN,cEMANet,cOCNet}
usually use self-attention to calculate the dot-product similarity in spatial space to encourage similar pixels to have a compact distance implicitly.
For example, the self-attention in~\cite{cNonLocal} implicitly pushed the feature representation of pixels belonging to the same class to be more similar to each other than those of pixels belonging to other classes. 
In our work, we devise a simple \emph{intra-class center-to-pixel loss} to guide the training, which can achieve this goal very effectively and produce improved accuracy.

\subsubsection{Intra-class Center-to-pixel Loss.} \label{sec:intra-c2p}
We define a simple but effective intra-class center-to-pixel loss to suppress the intra-class feature variance by penalizing large distance between a pixel feature and its class center. 
The Intra-class Center-to-pixel Loss $\mathcal{L}_{\text{intra-c2p}}$ is defined by:
\begin{equation}
    \small
    \mathcal{L}_{\text{intra-c2p}} = f_{\text{mse}}(\mathcal{D}_{\text{intra-c2p}}),
    \label{eq:CAR:intra_loss}
\end{equation}
where
\begin{equation}
    \small
    \mathcal{D}_{\text{intra-c2p}} = (1 - \mathbf{\sigma})\lvert \mathbf{Y}_{\text{flat}}\cdot\boldsymbol{\mu} - \mathbf{X}_\text{flat}\rvert.
    \label{eq:CAR:intra_diff} 
\end{equation}
In Eq.~(\ref{eq:CAR:intra_diff}), 
$\mathbf{\sigma}$ is a spatial mask indicating pixels being ignored (\textit{i.e.}, ignore label), 
$\mathbf{Y}_{\text{flat}}\cdot\boldsymbol{\mu}$ distributes the class centers $\boldsymbol{\mu}$ to the corresponding positions in each image.
Thus, our intra-class loss $\mathcal{L}_{\text{intra-c2p}}$
will push the pixel representations to their corresponding class center, using mean squared error (MSE) in Eq.~\eqref{eq:CAR:intra_diff}.

\subsection{Maximizing Inter-class Separation}

\subsubsection{Motivation.} 
Humans can robustly recognize an object by itself regardless which other objects it appears with. 
Conversely, if a classifier \emph{heavily} relies on the information from other classes to determine the classification result, 
it will easily produce wrong classification results when a rather rare class combination appears during inference.
Maximizing inter-class separation, or in another words, reducing the inter-class dependency, can therefore help the network generalize better, especially when the training set is small or is biased.
As shown in Fig.~\ref{fig:CAR:Intro}, the dog and sheep are mis-classified as the cow because cow and grass appear together more often in the training set. 
To improve the robustness of the model, we propose to reduce this inter-class dependency. 
To this end, the following two loss functions are defined.

\subsubsection{Inter-class Center-to-center Loss.}
\label{sec:inter_c2c}
The first loss function is to maximize the distance between any two different class centers.
Inspired by the center loss used in face recognition~\cite{cCenterLoss}, we propose to reduce the similarity between class centers $\boldsymbol{\mu}$,
which are the averaged features of each class calculated according to the GT mask. 
The \emph{inter-class relation} is defined by the dot-product similarity~\cite{cAttentionIsAllYourNeed} between any two classes as:
\begin{equation}
    \footnotesize
    \mathbf{A}_{\text{c2c}} = \text{softmax}(\frac{\boldsymbol{\mu}^{T}\cdot \boldsymbol{\mu}}{\sqrt{C}}), \:\:\ \mathbf{A}_{\text{c2c}} \in \mathbb{R}^{N_{class} \times N_{class}}.
    \label{eq:CAR:class-center-dot-sim}
\end{equation}

Moreover, since we only need to constrain the inter-class distance, only the non-diagonal elements are retained for the later loss calculation as: 
\begin{equation}
    \mathbf{D}_{\text{inter-c2c}} = \Big(1 - eye(N_{class})\Big)\mathbf{A}_{\text{c2c}}.
    \label{eq:CAR:class-direct-false-sim}
\end{equation}

We only penalize larger similarity values between any two different classes than a pre-defined threshold $ \frac{\epsilon_0}{ N_{class} - 1} $, \textit{i.e.},
\begin{equation}
    \mathcal{D}_{\text{inter-c2c}} = f_{\text{sum}}\Big(\text{max}(\mathbf{D}_{\text{inter-c2c}} - \frac{\epsilon_0}{ N_{class} - 1}, 0)\Big).
    \label{eq:CAR:inter-c2c}
\end{equation}
Thus, the Inter-class Center-to-center Loss $\mathcal{L}_{\text{inter-c2c}}$ is defined by:
\begin{equation}
    \mathcal{L}_{\text{inter-c2c}} = f_{\text{mse}}(\mathcal{D}_{\text{inter-c2c}}).
    \label{eq:CAR:class-center-loss}
\end{equation}
Here, a small margin is used in consideration of the feature space size and the mislabeled ground truth.

\subsubsection{Inter-class Center-to-pixel Loss.}
\label{sec:inter_c2p}
Maximizing only the distances between class centers does not necessarily result in separable representation for every individual pixels.
We further maximize the distance between a class center and any pixel that does not belong to this class.
More concretely, we first compute the center-to-pixel dot product as:
\begin{equation}
    \mathbf{\Lambda}_{\text{c2p}} = \boldsymbol{\mu}^{T}\cdot \mathbf{X_\text{flat}}, \:\:\ \mathbf{\Lambda}_{\text{c2p}} \in \mathbb{R}^{HW\times N_\text{class}}.
\end{equation}

Ideally, with the previous loss $\mathcal{L}_{\text{inter-c2c}}$, the features of all pixels belonging to the same class should be equal to that of the class center.
Therefore, we replace the intra-class dot product with its ideal value, namely using the class center $\boldsymbol{\mu}$ for calculating the intra-class dot product as:
\begin{equation}
    \mathbf{\Lambda}_c =diag(\boldsymbol{\mu}^{T} \cdot \boldsymbol{\mu}),
    \label{eq:c2c-dot-product}
\end{equation}
and the replacement effect is achieved by using masks as: 
\begin{equation}
    \mathbf{\Lambda^\prime} = \mathbf{\Lambda}_{\text{c2p}}(1 - \mathbf{Y}_{\text{flat}}) + \mathbf{\Lambda}_c.
    \label{eq:replacement}
\end{equation}

This updated dot product $\mathbf{\Lambda^\prime}$ is then used to calculate similarity across class axis with a softmax as: 
\begin{equation}
    \mathbf{A}_{\text{c2p}} = \text{softmax}(\mathbf{\Lambda^\prime}), \:\:\ \mathbf{A}_{\text{c2p}} \in \mathbb{R}^{HW\times N_\text{class}}.
    \label{eq:inter-c2p-similarity}
\end{equation}

Similar to the calculation of $\mathcal{L}_{\text{inter-c2c}}$ in the previous subsection, we have
\begin{equation}
    \mathbf{D}_{\text{inter-c2p}} = (1 - \mathbf{Y}_{\text{flat}})\mathbf{A}_{\text{c2p}},
\end{equation}
\begin{equation}
    \mathcal{D}_{\text{inter-c2p}} = f_{\text{sum}}\Big(\text{max}(\mathbf{D}_{\text{inter-c2p}} - \frac{\epsilon_1}{ N_\text{class} - 1}, 0)\Big).
    \label{eq:CAR:inter_c2p}
\end{equation}
Thus, the Inter-class Center-to-pixel Loss $\mathcal{L}_{\text{inter-c2p}}$ is defined by:
\begin{equation}
    \mathcal{L}_{\text{inter-c2p}} = f_{\text{mse}}(\mathcal{D}_{\text{inter-c2p}}).
    \label{eq:CAR:class-Pixel-loss}
\end{equation}

\subsection{Differences with OCR, ACFNet, CPNet, and CIPC}
Methods that are closely related to ours are OCR~\cite{cOCR}, ACFNet~\cite{cACFNet} and CPNet~\cite{cCPN}, which all focus on better utilizing class-level features and differ on how to extract the class centers and context features.
However, they all use a \textbf{simple concatenation} to fuse the original pixel feature and the complementary context feature.
For example, OCR and ACFNet first produce a coarse segmentation, which is supervised by the GT mask with a categorical cross-entropy loss, and then use this predicted coarse mask to generate the (soft) class centers by weighted summing all the pixel features.
OCR then aggregates these class centers according to their similarity to the original pixel feature termed as ``pixel-region relation'', resulting in a ``contextual feature''. 
Slightly differently from OCR, ACFNet directly uses the probability (from the predicted coarse mask) to aggregate class center, obtaining a similar context feature termed as ``attentional class feature''.
CPNet defines an affinity map, which is a binary map indicating if two spatial locations belong to the same class. Then, they use a sub-network to predict their ideal affinity map and use the soft version affinity map termed as ``Context Prior Map'' for feature aggregation, obtaining a class feature (center) and a context feature. Note that CPNet concatenates class feature, which is the updated pixel feature, and the context feature.

We also propose to utilize class-level contextual features.
Instead of extracting and fusing pixel features with sub-networks, we propose three loss functions to directly regularize training and encourage the learned features to maintain certain desired properties. The approach is simple but more effective thanks to the direct supervision (validated in Tab.~\ref{tab:CAR:AblationsBaseline}).
Moreover, our class center estimate is more accurate because we use the GT mask.
This strategy largely reduces the complexity of the network and introduces no computational overhead during inference.
Furthermore, it is compatible with all existing methods, including OCR, ACFNet and CPNet, demonstrating great generalization capability.

We also notice that Cross-Image Pixel Contrast (CIPC)~\cite{cCIPC} shares a similar high-level goal as our CAR, 
\textit{i.e.}, learning more similar representations for pixels belonging to the same class than to a different class. 
However, the approaches of achieving this goal are very different.
CIPC is motivated by contrastive learning while our CAR is motivated by the compositionality of the scene, for better generalization in the cases of rare class combinations.
Therefore, CIPC adopts the \emph{one-vs-rest} style InfoNCE loss, including the typical pixel-to-pixel loss and a special pixel-to-center loss.
In contrast,
\textbf{(1)} we propose an additional \emph{center-to-center} loss 
to regularize the inter-class dependency explicitly and effectively (see Table~\ref{tabCARAblationsParts}); 
\textbf{(2)} we use \emph{one-vs-one} style inter-class losses while CIPC uses \emph{one-vs-rest} style NCE loss. Compared to our \emph{one-vs-one} loss, using \emph{one-vs-rest} loss for training does not necessarily result in small inter-class similarity between the current class and every individual ``other'' classes and may increase the inter-class similarity among those ``other'' classes.
\textbf{(3)} we leave margins to prevent CAR regularizations, 
which is not the primary task of pixel classification,
from dominating the learning process.

\section{Experiments}

\subsection{Implementation} \label{sec:CAR:implementation}

\myparagraph{Training Settings.} 
For both CAR and baselines, we apply the settings common to most  works~\cite{cENCNet,cCFNet,cEMANet,cCCNet,cANNN}, including SyncBatchNorm, batch size = 16, weight decay (0.001), 0.01 initial LR, and poly learning decay with SGD during training. 
In addition, for the CNN backbones (\textit{e.g.}, ResNet), we set \textit{output stride} = 8 (see~\cite{cDeepLabV3}). 
Training iteration is set to 30k iterations unless otherwise specified. For the thresholds in Eq.~\ref{eq:CAR:inter-c2c} and Eq.~\ref{eq:CAR:inter_c2p}, we set $\epsilon_0 = 0.5$ and $\epsilon_1 = 0.25$.

\myparagraph{Determinism and Reproducibility}
Our implementations are based on the latest NVIDIA deterministic framework (2022), which means exactly the same results can be always reproduced with the same hardware and same training settings (including random seed).
To demonstrate the effectiveness of our CAR with equal comparisons, we reproduced all the baselines that we compare, all conducted with exactly the same settings unless otherwise specified. 

\subsection{Experiments on Pascal Context}

The Pascal Context~\cite{cPascalContext} dataset is split into 4,998/5,105 for training/test set.
We use its 59 semantic classes following the common practice~\cite{cOCR,cCFNet}.
Unless otherwise specified, both baselines and CAR are trained on the training set with 30k iterations. The ablation studies are presented in Sect.~\ref{sec:ablation_on_pascal}.

\subsubsection{Ablation Studies on Pascal Context} \label{sec:ablation_on_pascal}

\begin{table}[t]
\centering
\scriptsize
\caption{Ablation studies of adding CAR to different methods on Pascal Context dataset. All results are obtained with single scale test without flip.
``A'' means replacing the $3\times3 $ conv with $1\times1 $ conv (detailed in Sec.~\ref{sec:ablation_on_pascal}).
CAR improves the performance of different types of backbones (CNN \& Transformer) and head blocks (SA \& Uper), showing the generalizability of the proposed CAR.
}
\begin{tabular}{c|l|c|cc|c|c}
	\toprule[1pt]
	\rule{0pt}{2ex} & Methods         
	& $\mathcal{L}_{\text{intra-c2p}}$
	& $\mathcal{L}_{\text{inter-c2c}}$  
	& $\mathcal{L}_{\text{inter-c2p}}$
	& \quad A \quad\quad
	& mIOU (\%) \\
	\midrule[0.5pt]
	\midrule[0.5pt]
	R1&ResNet-50 + Self-Attention~\cite{cNonLocal} & -    & -    &    &   & 48.32 \\
	R2&  &               &               &               &\checkmark &48.56\\
	\midrule[0.5pt]
	R3&+ CAR & \checkmark    &               &               &           &49.17\\
    R4&   & \checkmark    & \checkmark    &               &           &49.79  \\
    R5&     & \checkmark    & \checkmark    & \checkmark    &           &50.01 \\
    R6&      & \checkmark    &               &               &\checkmark &49.62\\
    R7&       & \checkmark    & \checkmark    &               &\checkmark &50.00 \\
    R8&     & \checkmark    & \checkmark    & \checkmark    &\checkmark &\textbf{50.50}\\
    \midrule[0.5pt]
    \midrule[0.5pt]
    S1&Swin-Tiny + UperNet~\cite{cUper} & -    & -    &    &   & 49.62 \\
    S2&     &               &               &               &\checkmark &49.82\\
	\midrule[0.5pt]
	S3& + CAR & \checkmark    &               &               &           &49.01 \\
    S4&       & \checkmark    & \checkmark    &               &           &50.63  \\
    S5&      & \checkmark    & \checkmark    & \checkmark    &           &50.26 \\
    S6&      & \checkmark    &               &               &\checkmark & 49.62\\
    S7&       & \checkmark    & \checkmark    &               &\checkmark & 50.58  \\
    S8&      & \checkmark    & \checkmark    & \checkmark    &\checkmark & \textbf{50.78}\\
	\bottomrule[0.5pt]
\end{tabular}

\label{tabCARAblationsParts}
\end{table}

\myparagraph{CAR on ResNet-50 + Self-Attention.}
We firstly test the CAR with ``ResNet-50 + Self-Attention'' (w/o image-level block in~\cite{cCFNet}) to verify the effectiveness of the proposed loss functions, \textit{i.e.}, $\mathcal{L}_{\text{intra-c2p}}$, $\mathcal{L}_{\text{inter-c2c}}$, and $\mathcal{L}_{\text{inter-c2p}}$. 

As shown in Tab.~\ref{tabCARAblationsParts}, using $\mathcal{L}_{\text{intra-c2p}}$ directly improves 1.30 mIOU (48.32 vs 49.62); 
Introducing $\mathcal{L}_{\text{inter-c2c}}$ and $\mathcal{L}_{\text{inter-c2p}}$ further improves 0.38 mIOU and 0.50 mIOU; 
Finally, with all three loss functions,
the proposed CAR improves 2.18 mIOU from the regular ResNet-50 + Self-attention (48.32 vs 50.50).

\myparagraph{CAR on Swin-Tiny + Uper.}
``Swin-Tiny + Uper'' is a totally different architecture from ``ResNet-50 + Self-Attention~\cite{cNonLocal}''.
Swin~\cite{cSwin} is a recent Transformer-based backbone network.  
Uper~\cite{cUper} is based on the pyramid pooling modules (PPM)~\cite{cPSPNet} and FPN~\cite{cFPN}, focusing on extracting multi-scale context information. 
Similarly, as shown in Tab.~\ref{tabCARAblationsParts}, after adding CAR, the performance of Swin-Tiny + Uper also increases by 1.16, which shows our CAR can generalize to different architectures well.

\myparagraph{The Devil is In the Architecture's Detail.} 
We find it important to replace the leading $3\times3 $ conv (in the original method) with $1\times1 $ conv (Fig.~\ref{fig:CAR:Arch}B).
For example, $\mathcal{L}_{\text{intra-c2p}}$ and $\mathcal{L}_{\text{inter-c2p}}$ did not improve the performance in Swin-Tiny + Uper (Row S3 vs S1, and S5 vs S4 in Tab.~\ref{tabCARAblationsParts}).
A possible reason is that the network is trained to maximize the separation between different classes. However, if the two pixels lie on different sides of the segmentation boundary, a $3\times3 $ conv will merge the pixel representations from different classes, making the proposed losses harder to optimize.

To keep simplicity and maximize generalization, we use the same network configurations in our \textbf{all} experiments.
However, performance may be further improved with some minor dedicated modifications for each baseline when deploying our CAR.
For example, decreasing the filter number to 256 for the last conv layer of ResNet-50 + Self-Attention + CAR results in a further improvement to 51.00 mIOU (from 50.50).
Replacing the conv layer after PPM (inside Uper block, A3 in Fig.~\ref{fig:CAR:Arch}) from $3\times3$ to $1\times1$
in Swin-Tiny + UperNet boosts Swin (tiny \& large) + UperNet + CAR by an extra 0.5-1.0 mIOU.
%
We intentionally did \emph{not} exhaustively search these variants and \emph{not} report these results in any table since they did not generalize.

\begin{table}[h!]
\centering
\scriptsize
\caption{Ablation studies of adding CAR to different baselines on Pascal Context~\cite{cPascalContext} and COCOStuff-10K~\cite{cCocoStuff}. 
We deterministically reproduced all the baselines with the same settings.
All results are single-scale without flipping. 
CAR works very well in most existing methods.
\textbf{\S} means reducing the class-level threshold to 0.25 from 0.5. We found it is sensitive for some model variants to handle a large number of class.
Affinity loss~\cite{cCPN} and Auxiliary loss~\cite{cPSPNet} are applied on CPNet and OCR, respectively, since they highly rely on those losses.
}
\begin{tabular}{l|l|l|l|l}
	\toprule[1pt]
	\rule{0pt}{2ex} Methods & Backbone & Reference \quad\quad & \multicolumn{2}{c}{mIOU(\%)}\\
	\rule{0pt}{2ex}         &     &     & Pascal Context & COCO-Stuff10K  \\
	\midrule[0.5pt]
	\midrule[0.5pt]
    FCN~\cite{cFCN} & ResNet-50~\cite{cResnet}      &  CVPR2015 & 47.72 & 34.10  \\
	FCN + CAR & ResNet-50~\cite{cResnet}            &&48.40(\textcolor{blue}{+0.68}) &34.91(\textcolor{blue}{+0.81})\S  \\
	\midrule[0.1pt]
    FCN~\cite{cFCN} & ResNet-101~\cite{cResnet}     &  CVPR2015 & 50.93 &35.93  \\
	FCN + CAR & ResNet-101~\cite{cResnet}           &&51.39(\textcolor{blue}{+0.49}) &36.88(\textcolor{blue}{+0.95})\S  \\
	\midrule[0.5pt]
	DeepLabV3~\cite{cDeepLabV3Plus} & ResNet-50~\cite{cResnet} & ECCV2018 & 48.59 &34.96  \\
	DeepLabV3 + CAR & ResNet-50~\cite{cResnet}      && 49.53(\textcolor{blue}{+0.94}) &35.13(\textcolor{blue}{+0.17})  \\
	\midrule[0.1pt]
	DeepLabV3~\cite{cDeepLabV3Plus} & ResNet-101~\cite{cResnet} & ECCV2018 & 51.69 &36.92  \\
	DeepLabV3 + CAR & ResNet-101~\cite{cResnet}     & & 52.58(\textcolor{blue}{+0.89}) &37.39(\textcolor{blue}{+0.47})  \\
	\midrule[0.5pt]
	Self-Attention~\cite{cNonLocal} & ResNet-50~\cite{cResnet} & CVPR2018 & 48.32 & 34.35  \\
	Self-Attention + CAR & ResNet-50~\cite{cResnet} & & 50.50(\textcolor{blue}{+2.18}) &36.58(\textcolor{blue}{+2.23})\S  \\
	\midrule[0.1pt]
	Self-Attention~\cite{cNonLocal} & ResNet-101~\cite{cResnet} & CVPR2018 &51.59 & 36.53  \\
	Self-Attention + CAR & ResNet-101~\cite{cResnet} && 52.49(\textcolor{blue}{+0.9}) &38.15(\textcolor{blue}{+1.62})  \\
	\midrule[0.5pt]
	CCNet~\cite{cCCNet} & ResNet-50~\cite{cResnet}  & ICCV2019 & 49.15 &35.10  \\
	CCNet + CAR & ResNet-50~\cite{cResnet}          & & 49.56(\textcolor{blue}{+0.41}) &36.39(\textcolor{blue}{+1.29})  \\
	\midrule[0.1pt]
	CCNet~\cite{cCCNet} & ResNet-101~\cite{cResnet} & ICCV2019 & 51.41 &36.88  \\
	CCNet + CAR & ResNet-101~\cite{cResnet}         && 51.97(\textcolor{blue}{+0.56}) &37.56(\textcolor{blue}{+0.68})  \\
	\midrule[0.5pt]
	DANet~\cite{cDualAttention} & ResNet-101~\cite{cResnet} & CVPR2019& 51.45 &35.80  \\
	DANet + CAR & ResNet-101~\cite{cResnet}         && 52.57(\textcolor{blue}{+1.12}) &37.47(\textcolor{blue}{+1.67})  \\
	\midrule[0.5pt]
	CPNet~\cite{cCPN} & ResNet-101~\cite{cResnet} & CVPR2020& 51.29&36.92  \\
	CPNet + CAR & ResNet-101~\cite{cResnet}         && 51.98(\textcolor{blue}{+0.69})&37.12(\textcolor{blue}{+0.20})\S  \\
	\midrule[0.5pt]
	OCR~\cite{cOCR} & HRNet-W48~\cite{cHRNet}       & ECCV2020& 54.37 & 38.22 \\
	OCR + CAR & HRNet-W48~\cite{cHRNet}             && 54.99(\textcolor{blue}{+0.62})  & 39.53(\textcolor{blue}{+1.31}) \\
	\midrule[0.5pt]
	UperNet~\cite{cUper} & Swin-Tiny~\cite{cSwin}   & ICCV2021& 49.62 & 36.07  \\
	UperNet + CAR & Swin-Tiny~\cite{cSwin}          && 50.78(\textcolor{blue}{+1.16})  & 36.63(\textcolor{blue}{+0.56})\S  \\
	\midrule[0.1pt]
	UperNet~\cite{cUper} & Swin-Large~\cite{cSwin}  & ICCV2021 & 57.48 &44.25  \\ 
	UperNet + CAR & Swin-Large~\cite{cSwin}         & & 58.97(\textcolor{blue}{+1.49}) & 44.88(\textcolor{blue}{+0.63})  \\
	\midrule[0.1pt]
	CAA~\cite{cCAA} & EfficientNet-B5~\cite{cEfficientNet}& AAAI2022& 57.79  & 43.40  \\ 
	CAA + CAR & EfficientNet-B5~\cite{cEfficientNet} & & 58.96(\textcolor{blue}{+1.17})  & 43.93(\textcolor{blue}{+0.53})  \\
	\bottomrule[1pt]
\end{tabular}
\label{tab:CAR:AblationsBaseline}
\end{table}

\myparagraph{CAR on Different Baselines.}
After we have verified the effectiveness of each part of the proposed CAR, we then tested CAR on multiple well-known baselines. All of the baselines were reproduced under similar conditions (see Sect.~\ref{sec:CAR:implementation}). 
Experimental results shown in Tab.~\ref{tab:CAR:AblationsBaseline} demonstrate the generalizability of our CAR on different backbones and methods.

\begin{figure}[tb]
\centering
\includegraphics[width=1.0\linewidth]{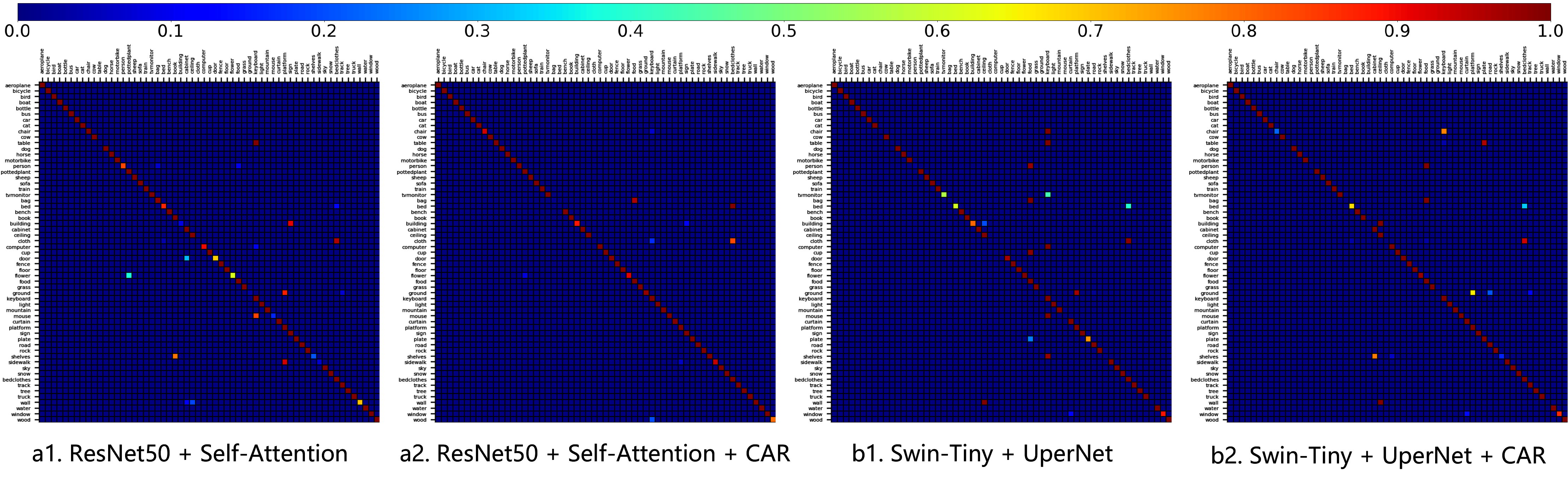}
\caption{Class dependency maps generated on Pascal Context test set. 
One may zoom in to see class names. 
A hotter color means that the class has higher dependency to the corresponding class, and vice versa. It is obvious that our CAR reduces the inter-class dependency, thus providing better generalizability (see Figs.~\ref{fig:CAR:Intro} and~\ref{fig:CAR:PixelRelationMap}).
}
\label{fig:CAR:ClassDependencyMap}
\end{figure}

\begin{figure}[t]
\centering
\begin{subfigure}[b]{0.49\textwidth}
     \centering
     \includegraphics[width=\textwidth]{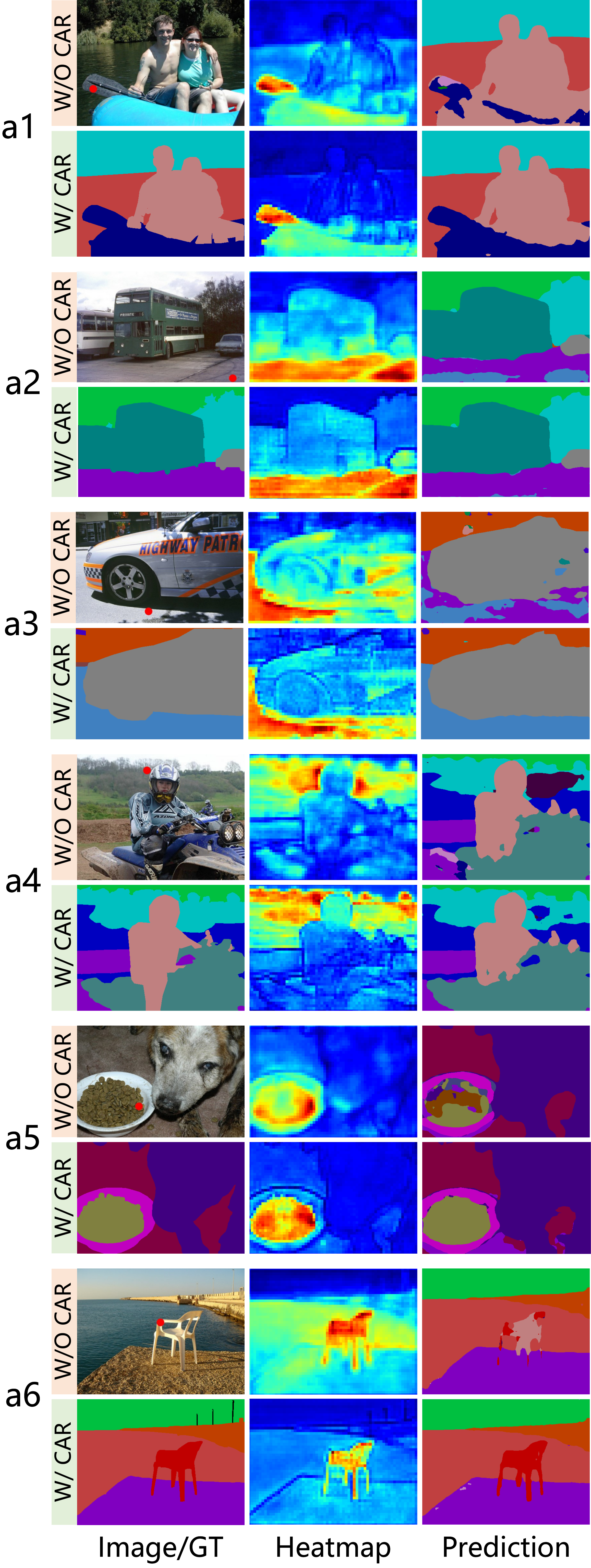}
     \caption{ResNet50 + Self-Attention}
     \label{fig:CAR:PixelRelationMap:ResNet}
 \end{subfigure}
 \hfill
 \begin{subfigure}[b]{0.49\textwidth}
     \centering
     \includegraphics[width=\textwidth]{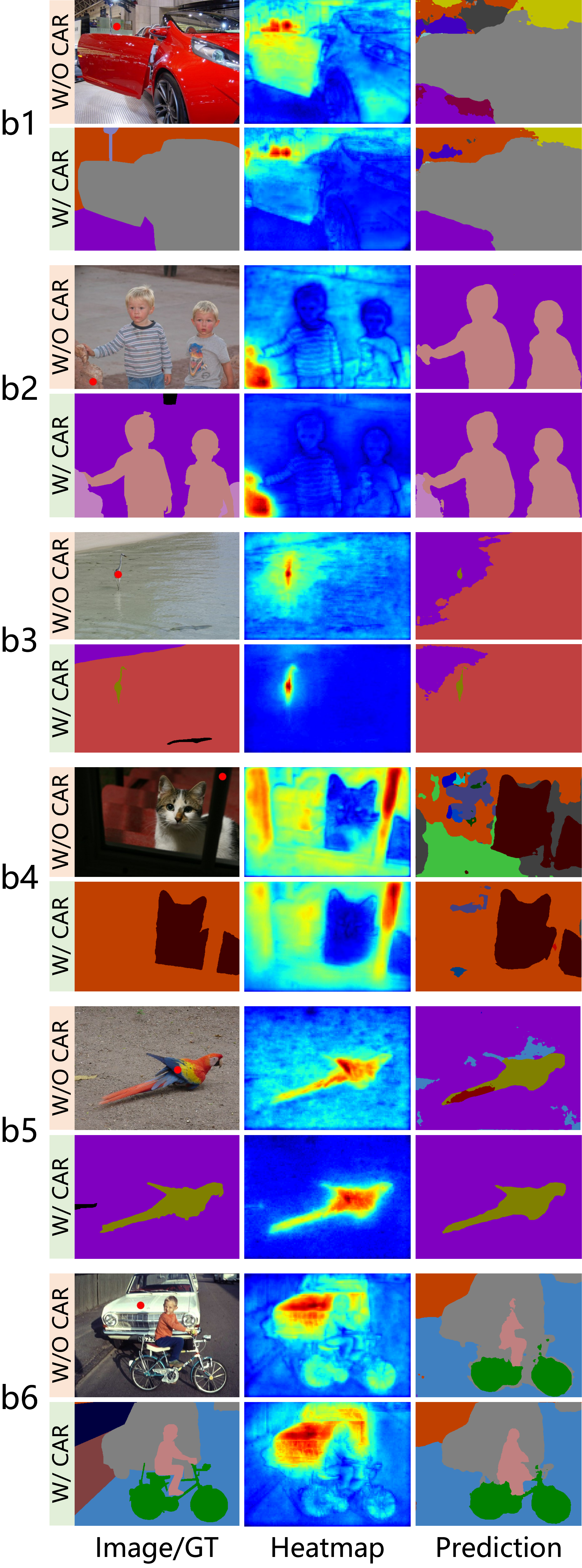}
     \caption{Swin-Tiny + UperNet}
     \label{fig:CAR:PixelRelationMap:SwinTiny}
 \end{subfigure}
 \hfill
\caption{
Visualization of the feature similarity between a given pixel (marked with a red dot in the image) and all pixels,
as well as the segmentation results on Pascal Context test set. 
A hotter color denotes larger similarity value. 
Apparently, our CAR reduces the inter-class dependency and exhibits better generalization ability,
where energies are better restrained in the intra-class pixels.
}
\label{fig:CAR:PixelRelationMap}
\end{figure}

\myparagraph{Visualization of Class Dependency Maps.} 
In Fig.~\ref{fig:CAR:ClassDependencyMap}, we present the class dependency maps calculated on the complete Pascal Context \emph{test} set, where every pixel stores the dot-product similarities between every two class centers. 
The maps indicate the inter-class dependency obtained with the standard ResNet-50 + Self-Attention and Swin-Tiny + UperNet, and the effect of applying our CAR.  
A hotter color means that the class has higher dependency on the corresponding class, and vice versa. 
According to Fig.~\ref{fig:CAR:ClassDependencyMap} a1-a2, we can easily observe that the inter-class dependency has been significantly reduced with CAR on ResNet50 + Self-Attention. 
Fig.~\ref{fig:CAR:ClassDependencyMap} b1-b2 show a similar trend when tested with different backbones and head blocks.
This partially explains the reason why baselines with CAR generalize better on rarely seen class combinations (Figs.~\ref{fig:CAR:Intro} and~\ref{fig:CAR:PixelRelationMap}).
Interestingly, we find that the class-dependency issue is more serious in Swin-Tiny + Uper, but our CAR can still reduce its dependency level significantly.

\myparagraph{Visualization of Pixel-relation Maps.}
In Fig.~\ref{fig:CAR:PixelRelationMap},
we visualize the pixel-to-pixel relation energy map, based on the dot-product similarity between a red-dot marked pixel and other pixels, as well as the predicted results for different methods, for comparison.
Examples are from Pascal Context test set.
As we can see, with CAR supervision,
the existing models focus better on objects themselves rather than other objects. Therefore, this reduces the possibility of the classification errors because of the class-dependency bias.

\subsection{Experiments on COCOStuff-10K}

COCOStuff-10K dataset~\cite{cCocoStuff} is widely used for evaluating the robustness of semantic segmentation models~\cite{cEMANet,cOCR}. The COCOStuff-10k dataset is a very challenging dataset containing 171 labeled classes and 9000/1000 images for training/test.
As shown in Tab.~\ref{tab:CAR:AblationsBaseline}, all of the tested baselines gain performance boost ranging from 0.17\% to 2.23\% with our proposed CAR.
This demonstrates the generalization ability of our CAR when handling a large number of classes.

\section{Conclusions and Future Work}

In this paper, we have aimed to make a better use of class level context information.
We have proposed a universal class-aware regularizations (CAR) approach to regularize the training process and boost the differentiability of the learned pixel representations.
To this end, we have proposed to minimize the intra-class feature variance and maximize the inter-class separation simultaneously. 
Experiments conducted on benchmark datasets with extensive ablation studies have validated the effectiveness of the proposed CAR approach, which has boosted the existing models' performance by up to 2.18\% mIOU on Pascal Context and 2.23\% on COCOStuff-10k with no extra inference overhead. \\

\noindent\textbf{Acknowledgement} This research depends on the NVIDIA determinism framework. We appreciate the support from @duncanriach and @reedwm at NVIDIA and TensorFlow team. 

\clearpage

%
%
\bibliographystyle{splncs04}
\bibliography{egbib}

\begin{thebibliography}{10}
\providecommand{\url}[1]{\texttt{#1}}
\providecommand{\urlprefix}{URL }
\providecommand{\doi}[1]{https://doi.org/#1}

\bibitem{cCocoStuff}
Caesar, H., Uijlings, J., Ferrari, V.: {COCO-Stuff: Thing and Stuff Classes in
  Context}. In: IEEE Conference on Computer Vision and Pattern Recognition
  (2018)

\bibitem{cDeepLab}
Chen, L.C., Papandreou, G., Kokkinos, I., Murphy, K., Yuille, A.L.: Deeplab:
  Semantic image segmentation with deep convolutional nets, atrous convolution,
  and fully connected crfs. IEEE Transactions on Pattern Analysis and Machine
  Intelligence  (2017)

\bibitem{cDeepLabV3}
Chen, L.C., Papandreou, G., Schroff, F., Adam, H.: Rethinking atrous
  convolution for semantic image segmentation (2017)

\bibitem{cDeepLabV3Plus}
Chen, L.C., Zhu, Y., Papandreou, G., Schroff, F., Adam, H.: Encoder-decoder
  with atrous separable convolution for semantic image segmentation. In:
  European Conference on Computer Vision (2018)

\bibitem{cHANet}
Choi, S., Kim, J.T., Choo, J.: Cars can’t fly up in the sky: Improving
  urban-scene segmentation via height-driven attention networks. In: IEEE
  Conference on Computer Vision and Pattern Recognition (2020)

\bibitem{cViT}
Dosovitskiy, A., Beyer, L., Kolesnikov, A., Weissenborn, D., Zhai, X.,
  Unterthiner, T., Dehghani, M., Minderer, M., Heigold, G., Gelly, S.,
  Uszkoreit, J., Houlsby, N.: An image is worth 16x16 words: Transformers for
  image recognition at scale. In: International Conference on Learning
  Representations (2021)

\bibitem{cDualAttention}
Fu, J., Liu, J., Tian, H., Li, Y., Bao, Y., Fang, Z., Lu, H.: Dual attention
  network for scene segmentation. In: IEEE Conference on Computer Vision and
  Pattern Recognition (2019)

\bibitem{cResnet}
He, K., Zhang, X., Ren, S., Sun, J.: Deep residual learning for image
  recognition. In: IEEE Conference on Computer Vision and Pattern Recognition
  (2016)

\bibitem{cCAA}
Huang, Y., Kang, D., Jia, W., He, X., liu, L.: Channelized axial attention -
  considering channel relation within spatial attention for semantic
  segmentation. In: AAAI (2022)

\bibitem{cCCNet}
Huang, Z., Wang, X., Wei, Y., Huang, L., Shi, H., Liu, W., Huang, T.S.: Ccnet:
  Criss-cross attention for semantic segmentation. IEEE Transactions on Pattern
  Analysis and Machine Intelligence  (2020)

\bibitem{cEMANet}
Li, X., Zhong, Z., Wu, J., Yang, Y., Lin, Z., Liu, H.: Expectation-maximization
  attention networks for semantic segmentation. In: International Conference on
  Computer Vision (2019)

\bibitem{cFPN}
Lin, T.Y., Dollá, P., Girshick, R., He, K., Hariharan, B., Belongie, S.:
  Feature pyramid networks for object detection. In: IEEE Conference on
  Computer Vision and Pattern Recognition (2017)

\bibitem{cDependencyNet}
Liu, M., Schonfeld, D., Tang, W.: Exploit visual dependency relations for
  semantic segmentation. In: IEEE Conference on Computer Vision and Pattern
  Recognition (2021)

\bibitem{cSwin}
Liu, Z., Lin, Y., Cao, Y., Hu, H., Wei, Y., Zhang, Z., Lin, S., Guo, B.: Swin
  transformer: Hierarchical vision transformer using shifted windows. In: ICCV
  (2021)

\bibitem{cConvNeXT}
Liu, Z., Mao, H., Wu, C.Y., Feichtenhofer, C., Darrell, T., Xie, S.: A convnet
  for the 2020s. In: IEEE Conference on Computer Vision and Pattern Recognition
  (2022)

\bibitem{cFCN}
Long, J., Shelhamer, E., Darrell, T.: Fully convolutional networks for semantic
  segmentation. In: IEEE Conference on Computer Vision and Pattern Recognition
  (2015)

\bibitem{cCityScapes}
Marius, C., Mohamed, O., Sebastian, R., Timo, R., Markus, E., Rodrigo, B., Uwe,
  F., Roth, S., Bernt, S.: The cityscapes dataset for semantic urban scene
  understanding. In: IEEE Conference on Computer Vision and Pattern Recognition
  (2016)

\bibitem{cEfficientNet}
Mingxing, T., Quoc, L.: Efficientnet: Rethinking model scaling for
  convolutional neural networks. In: International Conference on Machine
  Learning (2019)

\bibitem{cPascalContext}
Mottaghi, R., Chen, X., Liu, X., Cho, N.G., Lee, S.W., Fidler, S., Urtasun, R.,
  Yuille, A.: The role of context for object detection and semantic
  segmentation in the wild. In: IEEE Conference on Computer Vision and Pattern
  Recognition (2014)

\bibitem{cDPT}
Ranftl, R., Bochkovskiy, A., Koltun, V.: Vision transformers for dense
  prediction. In: ICCV (2021)

\bibitem{cVGG}
Simonyan, K., Zisserman, A.: Very deep convolutional networks for large-scale
  image recognition. In: International Conference on Learning Representations
  (2015)

\bibitem{cSETR}
Sixiao, Z., Jiachen, L., Hengshuang, Z., Xiatian, Z., Zekun, L., Yabiao, W.,
  Yanwei, F., Jianfeng, F., Tao, X., H.S., T.P., Li, Z.: Rethinking semantic
  segmentation from a sequence-to-sequence perspective with transformers. In:
  IEEE Conference on Computer Vision and Pattern Recognition (2021)

\bibitem{cAttentionIsAllYourNeed}
Vaswani, A., Shazeer, N., Parmar, N., Uszkoreit, J., Jones, L., Gomez, A.N.,
  Łukasz Kaiser, Polosukhin, I.: Attention is all you need. In: Conference on
  Neural Information Processing Systems (2017)

\bibitem{cHRNet}
Wang, J., Sun, K., Cheng, T., Jiang, B., Deng, C., Zhao, Y., Liu, D., Mu, Y.,
  Tan, M., Wang, X., Liu, W., Xiao, B.: Deep high-resolution representation
  learning for visual recognition. IEEE Transactions on Pattern Analysis and
  Machine Intelligence  (2020)

\bibitem{cCIPC}
Wang, W., Zhou, T., Yu, F., Dai, J., Konukoglu, E., Van~Gool, L.: Exploring
  cross-image pixel contrast for semantic segmentation. In: ICCV. pp.
  7303--7313 (2021)

\bibitem{cNonLocal}
Wang, X., Girshick, R., Gupta, A., He, K.: Non-local neural networks. In: IEEE
  Conference on Computer Vision and Pattern Recognition (2018)

\bibitem{cCenterLoss}
Wen1, Y., Zhang, K., Li, Z., Qiao, Y.: Discriminative feature learning approach
  for deep face recognition. In: European Conference on Computer Vision (2016)

\bibitem{cFastFCN}
Wu, H., Zhang, J., Huang, K., Liang, K., Yizhou, Y.: Fastfcn: Rethinking
  dilated convolution in the backbone for semantic segmentation (2019)

\bibitem{cUper}
Xiao, T., Liu, Y., Zhou, B., Jiang, Y., Sun, J.: Unified perceptual parsing for
  scene understanding. In: European Conference on Computer Vision (2018)

\bibitem{cCPN}
Yu, C., Wang, J., Gao, C., Yu, G., Shen, C., Sang, N.: Context prior for scene
  segmentation. In: IEEE Conference on Computer Vision and Pattern Recognition
  (2020)

\bibitem{cOCR}
Yuan, Y., Chen, X., Wang, J.: Object-contextual representations for semantic
  segmentation. In: European Conference on Computer Vision (2020)

\bibitem{cOCNet}
Yuan, Y., Huang, L., Guo, J., Zhang, C., Chen, X., Wang, J.: Ocnet: Object
  context network for scene parsing. International Journal of Computer Vision
  (2021)

\bibitem{cACFNet}
Zhang, F., Chen, Y., Li, Z., Hong, Z., Liu, J., Ma, F., Han, J., Ding, E.:
  Acfnet: Attentional class feature network for semantic segmentation. In:
  International Conference on Computer Vision (2019)

\bibitem{cENCNet}
Zhang, H., Dana, K., Shi, J., Zhang, Z., Wang, X., Tyagi, A., Agrawal, A.:
  Context encoding for semantic segmentation. In: IEEE Conference on Computer
  Vision and Pattern Recognition (2018)

\bibitem{cCFNet}
Zhang, H., Zhan, H., Wang, C., Xie, J.: Semantic correlation promoted
  shape-variant context for segmentation. In: IEEE Conference on Computer
  Vision and Pattern Recognition (2019)

\bibitem{cPSPNet}
Zhao, H., Shi, J., Qi, X., Wang, X., Jia, J.: Pyramid scene parsing network.
  In: IEEE Conference on Computer Vision and Pattern Recognition (2017)

\bibitem{cANNN}
Zhu, Z., Xu, M., Bai, S., Huang, T., Bai, X.: Asymmetric non-local neural
  networks for semantic segmentation. In: International Conference on Computer
  Vision (2019)

\end{thebibliography}

\clearpage

\appendix

\section{Appendix}

\subsection{Extra technical details}

\subsubsection{Deterministic}

Control variables are very important for all scientific research. 
In computer vision, we always use the same backbones and the same datasets when verifying the difference between two methods.

Without using ``deterministic'' technology, all operations in neural networks contain some randomness.
Nowadays, with the latest deterministic technology and fixed seeds, experiments can be conducted in a fully-controlled environment. 
This means that the performance difference between different settings (\textit{i.e.}, w/ and w/o CAR) is not affected by this randomness any more but faithfully reflects the effectiveness of different methods.

In Tab.~\ref{tabCARRand}, we report the performance of our proposed CAR (ResNet-50 + Self-attention and Swin-Tiny + UperNet) with different seeds for readers who are interested in how our CAR performs when trained with different random seeds.
As it is shown, our CAR consistently improves the mIOU over its baseline using different random seeds, demonstrating the effectiveness of the CAR.

\begin{table}[h]
	\centering
	\small
	\caption{Ablation studies of our proposed CAR using different random seeds on the Pascal Context dataset.}
	\begin{tabular}{l|l|l|l}
		\toprule[1pt]
		\rule{0pt}{2ex} Methods &
		\multicolumn{3}{c}{Seed (mIOU\%)}\\
		& 0 (default) & 1 & 2\\
		\midrule[0.5pt]
		ResNet-50 + Self-Attention &48.32 &47.54 &47.69\\
		ResNet-50 + Self-Attention + CAR 
		& 50.50(\textcolor{blue}{+2.18}) 
		& 50.20(\textcolor{blue}{+2.66}) 
		& 50.59(\textcolor{blue}{+2.90})\\
		\midrule[0.5pt]
		Swin-Tiny + UperNet &49.62 &49.24 &49.54\\
		Swin-Tiny + UperNet + CAR 
		& 50.78(\textcolor{blue}{+1.16}) 
		& 50.57(\textcolor{blue}{+1.33}) 
		& 50.75(\textcolor{blue}{+1.21})\\
		\bottomrule[0.5pt]
	\end{tabular}
	\label{tabCARRand}
\end{table}

\subsection{Extra experiments}

\subsubsection{Ablation studies on batch class center}

In our experiments, we calculated the class centers using all the training images in a \emph{batch} to alleviate the negative impact of noisy images. 
Here, we investigate the impact of using the class center of each individual image for class-aware regularizations.

\begin{table}[h]
	\centering
	\small
	\caption{Comparison of mIOUs (\%) obtained when using batch class center vs image class center in CAR.}
	\begin{tabular}{c|c|c|c}
		\toprule[1pt]
		\rule{0pt}{2ex} Methods &
		\quad Baseline \quad\quad &
		\multicolumn{2}{c}{CAR} \\
		&&Image Class Center&Batch Class Center\\
		\midrule[0.5pt]
		ResNet-50 + Self-Attention &48.32 &49.78
		&50.50\\
		\midrule[0.5pt]
		Swin-Tiny + UperNet & 49.62 & 49.45
		&50.78\\
		\bottomrule[0.5pt]
	\end{tabular}
	\label{tabCarImageCenter}
\end{table}

\subsubsection{CAR using Moving Average.} We also implemented a moving average version of CAR which tracks the class center $\boldsymbol{\mu}$ with moving average similar to BatchNorm.
As shown in Tab.~\ref{tabCARMovingAverage}, we find this moving average version of CAR negatively impacts both ResNet-50 + Self-Attention and Swin-Tiny + Uper.

\begin{table}[t]
	\centering
	\small
	\caption{Ablation studies of adding moving average to CAR on Pascal Context. Decay rate stands for the effect of old class center.}
	\begin{tabular}{c|c|c|c|c}
		\toprule[1pt]
		\rule{0pt}{2ex} Methods &
		\quad CAR \quad\quad &
		\multicolumn{3}{c}{CAR (Moving Average)} \\
		&&0.8&0.9&0.99\\
		\midrule[0.5pt]
		ResNet-50 + Self-Attention &50.50 &49.80(\textcolor{red}{$-$0.70})
		&50.26(\textcolor{red}{$-$0.24})
		&49.96(\textcolor{red}{$-$0.54})\\
		\midrule[0.5pt]
		Swin-Tiny + UperNet &50.78 &49.56(\textcolor{red}{$-$1.22})
		&50.03(\textcolor{red}{$-$0.75})
		&48.93(\textcolor{red}{$-$1.85})\\
		\bottomrule[0.5pt]
	\end{tabular}
	\label{tabCARMovingAverage}
\end{table}

\subsubsection{CAR without intra-class loss.} Table~\ref{tabCARWOIntraClassLoss} below shows that, including only inter-c2c loss and inter-c2p still improves the result.

\begin{table}[t]
\centering
\small
\caption{Extra ablation studies for CAR without intra-class loss.}
\begin{tabular}{l|c|cc|c|c}
	\toprule[0.7pt]
	\rule{0pt}{2ex}        
	& $\mathcal{L}_{\text{intra-c2p}}$
	& $\mathcal{L}_{\text{inter-c2c}}$  
	& $\mathcal{L}_{\text{inter-c2p}}$
	& A \quad
	& mIOU
    \\
	\midrule[0.5pt]
	\midrule[0.5pt]
	ResNet-50 + Self-Attention & -    & -    &    &   & 48.32 \\
    &               &               &               &\checkmark & 48.56 \\
	\midrule[0.5pt]
	+ CAR      & & \checkmark    & \checkmark    &  & 49.31\\
                & & \checkmark    & \checkmark    &\checkmark & 50.23\\
	\bottomrule[0.7pt]
\end{tabular}
\label{tabCARWOIntraClassLoss}
\end{table}

\subsubsection{Exceeding state-of-the-art (SOTA) in Pascal Context}
\label{secCARBeyondSOTAPascalContext}

The main motivation of our CAR is to utilize class-level information as regularizations during training to boost the performance of all existing methods.
However, following the convention and also for readers who are interested, we compare with state-of-the-art methods in Tab.~\ref{tabCARBeyondSOTAPascalContext} regardless their architectures are related to ours or not.
%
Since Swin~\cite{cSwin} is not compatible with dilation, we use JPU~\cite{cFastFCN} as the substitution to obtain features with output stride = 8.  
Uper contains an FPN~\cite{cFPN} module that can obtain features with output stride = 4.

Boosted by our CAR, the strong model ConvNext-Large~\cite{cConvNeXT} + CAA~\cite{cCAA} achieved the performance of 62.70\% mIOU under single-scale testing, and 63.91\% under multi-scale testing. 
Also, we found increasing the training iterations from the default 30K to 40K when using Adam optimizer can further increase performance in Pascal Context dataset. Thus, the SOTA single model performance has now been boosted to 62.97\% under single-scale testing, and 64.12\% under multi-scale testing.
This has outperformed the previous SOTA single model, \textit{i.e.}, EfficientNetB7 + CAA, by a large margin.

\begin{table}[h]
\centering
\small
\caption{
Experiments on boosting the SOTA single-model performance on Pascal Context by our CAR.
See Sec.~\ref{secCARBeyondSOTAPascalContext} for the details. \textit{\S}: We report previous SOTA score as reference. \textit{SS}: mIOU on Single scale without flipping. \textit{MF}: Multi-scale with flipping. JPU is used to get features with output stride = 8. \textit{Aux}: Apply auxiliary loss during training, see~\cite{cPSPNet}. \textit{Iters}: training iterations. \textit{WP}:Linear learning rate warmup
}
\begin{tabular}{l|l|c|c|c|c|c|c}
	\toprule[1pt]
	\rule{0pt}{2ex} 
	Methods
	& Backbone
    & Aux
	& Optimizer
	& Iters
	& WP
	& SS(\%)
	& MF(\%) \\
    \midrule[0.5pt]
    CAA\S & EfficientNet-B7-D8 &\checkmark & SGD &30K&& - & 60.30\\
    \midrule[0.5pt]
    UperNet & Swin-Large && SGD &30K&&57.48 & 59.45 \\
    UperNet + CAR & Swin-Large && SGD &30K&& 58.97&60.76 \\
    \midrule[0.5pt]
    CAA & Swin-Large + JPU && SGD &30K&& 58.31& 59.75\\
    CAA + CAR & Swin-Large + JPU && SGD &30K&& 59.84&61.46\\
    CAA + CAR & Swin-Large + JPU && Adam &30K&& 60.68&62.21\\
    \midrule[0.5pt]
    CAA  & ConvNeXt-Large + JPU && SGD &30K&& 60.48 & 61.80\\
    CAA + CAR & ConvNeXt-Large + JPU&& SGD&30K&&61.40& 62.69\\
    CAA + CAR & ConvNeXt-Large + JPU&& Adam &30K&& 62.65&63.77\\
    CAA + CAR & ConvNeXt-Large + JPU&\checkmark& Adam &30K&&62.70&63.91\\
    CAA + CAR & ConvNeXt-Large + JPU&\checkmark& Adam &40K&&62.97&\textbf64.12\\
    CAA + CAR & ConvNeXt-Large + JPU&\checkmark& Adam &40K&\checkmark&63.13 & 64.17 \\
    \bottomrule
\end{tabular}
\label{tabCARBeyondSOTAPascalContext}
\end{table}

\subsubsection{Exceeding SOTA performance in COCOStuff-10K}
\label{secCARBeyondSOTACOCOStuff10k}

Simliar to Sec.~\ref{secCARBeyondSOTAPascalContext}, in Tab.~\ref{tabCARBeyondSOTACOCOStuff10k}, boosted by our CAR, the strong model ConvNext-Large~\cite{cConvNeXT} + CAA achieved the performance of 49.03\% mIOU under single-scale testing, and 50.01\% under multi-scale testing. 
This has also outperformed the previous SOTA single model, \textit{i.e.}, EfficientNetB7 + CAA, by a large margin.

\begin{table}[t]
\centering
\small
\caption{Experiments on boosting SOTA on COCOStuff10k, levering the previous single model SOTA and boosted by our CAR. See Sec.~\ref{secCARBeyondSOTACOCOStuff10k} for details. \textit{\S}: We report the original SOTA scores. \textit{SS}: Single scale without flipping. \textit{MF}: Multi-scale with flipping. \textit{Aux} Apply auxiliary loss during training, see~\cite{cPSPNet}.
}
\begin{tabular}{l|l|c|c|c|c}
	\toprule[1pt]
	\rule{0pt}{2ex} 
	Methods
	& Backbone
    & Aux
	& Optimizer
	& SS mIOU(\%)
	& MF mIOU(\%) \\
    \midrule[0.5pt]
    CAA\S & EfficientNet-B7-D8 &\checkmark & SGD & - & 45.40\\
    \midrule[0.5pt]
    UperNet & Swin-Large && SGD &44.25 & 46.10\\
    UperNet + CAR & Swin-Large && SGD & 44.88& 46.64\\
    \midrule[0.5pt]
    CAA & Swin-Large + JPU && SGD & 44.22&45.31 \\
    CAA + CAR & Swin-Large + JPU && SGD & 45.48 & 46.99 \\
    \midrule[0.5pt]
    CAA  & ConvNeXt-Large + JPU && SGD & 46.49 & 47.23\\
    CAA + CAR & ConvNeXt-Large + JPU&& SGD& 46.70& 47.77\\
    CAA + CAR & ConvNeXt-Large + JPU&& Adam & 48.20&48.83 \\
    CAA + CAR & ConvNeXt-Large + JPU&\checkmark& Adam &\textbf{49.03}&\textbf{50.01}\\
    \bottomrule
\end{tabular}
\label{tabCARBeyondSOTACOCOStuff10k}
\end{table}

\subsection{Extra Visualizations}

\subsubsection{Visualization of OCRNet in Pascal Context}

\begin{figure}[h]
\centering
\includegraphics[width=1.0\linewidth]{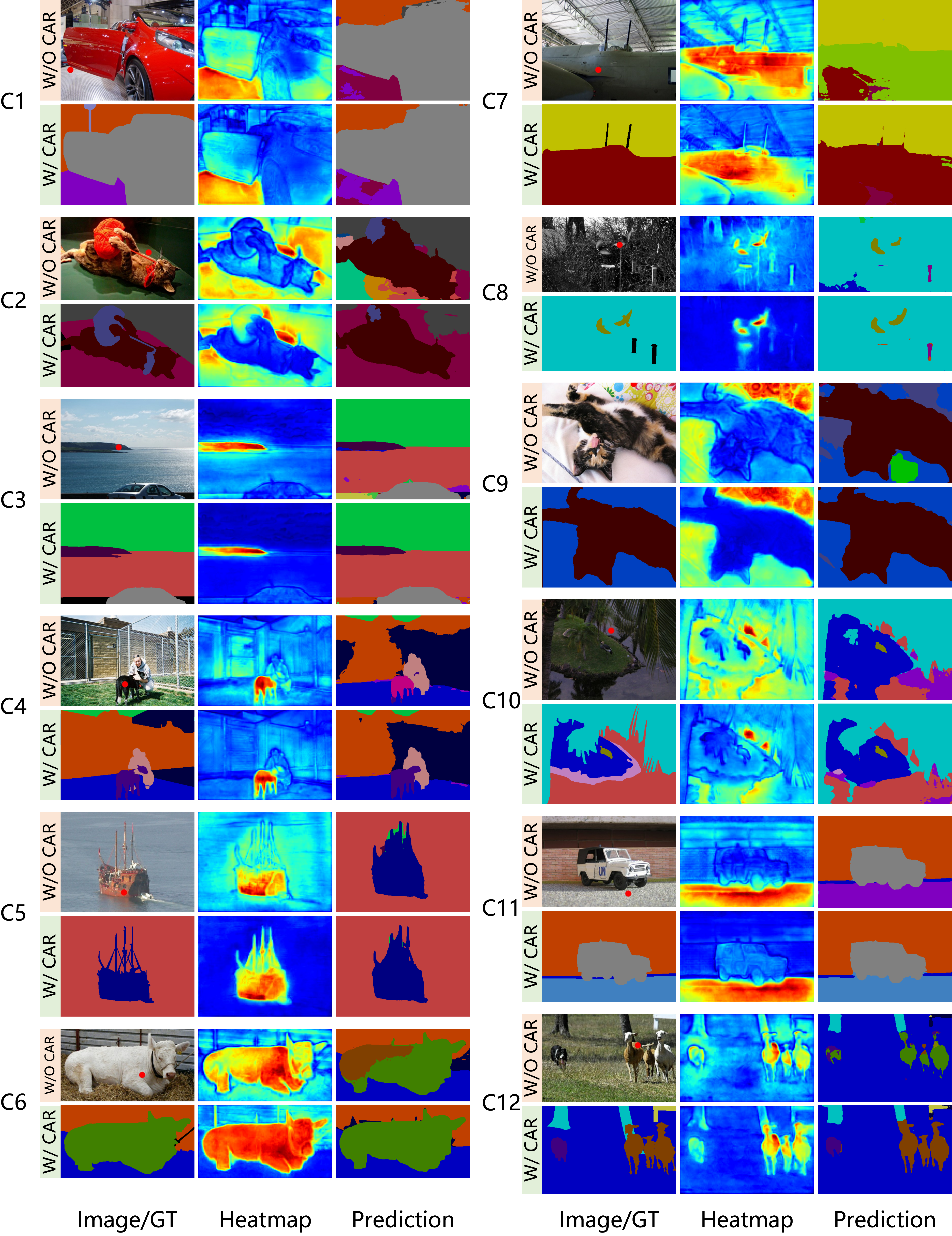}
\caption{Visualization of the feature similarity between a given pixel (marked with a red dot in the image) and all other pixels,
as well as the segmentation results of \textbf{HRNetW48~\cite{cHRNet} + OCR~\cite{cOCR}} on Pascal Context test set. 
A hotter color denotes a greater similarity value. 
}
\label{fig:CAR:PixelRelationMap:OCRHRNetW48}
\end{figure}

Similar to the main paper, in Fig.~\ref{fig:CAR:PixelRelationMap:OCRHRNetW48}, we visualize the pixel-to-pixel relation energy maps obtained with HRNetW48~\cite{cOCR} + OCR~\cite{cOCR}. This figure shows that our CAR can further improve the robustness of class center based models by making better use of the class center.
Interestingly, as shown in C12 of Fig.~\ref{fig:CAR:PixelRelationMap:OCRHRNetW48} and Fig. 1 shown in our main paper what is predicted by ResNet-50 + Self-Attention, we find cow/sheep/dog misclassification is a common issue in many semantic segmentation models, especially when \textit{i.e.} grass and cow co-exist frequently during training.
%
This issue is better addressed by our CAR due to its reduced inter-class dependency.

\subsubsection{Visualization of DeepLab in Pascal Context}

\begin{figure}[h]
\centering
\includegraphics[width=1.0\linewidth]{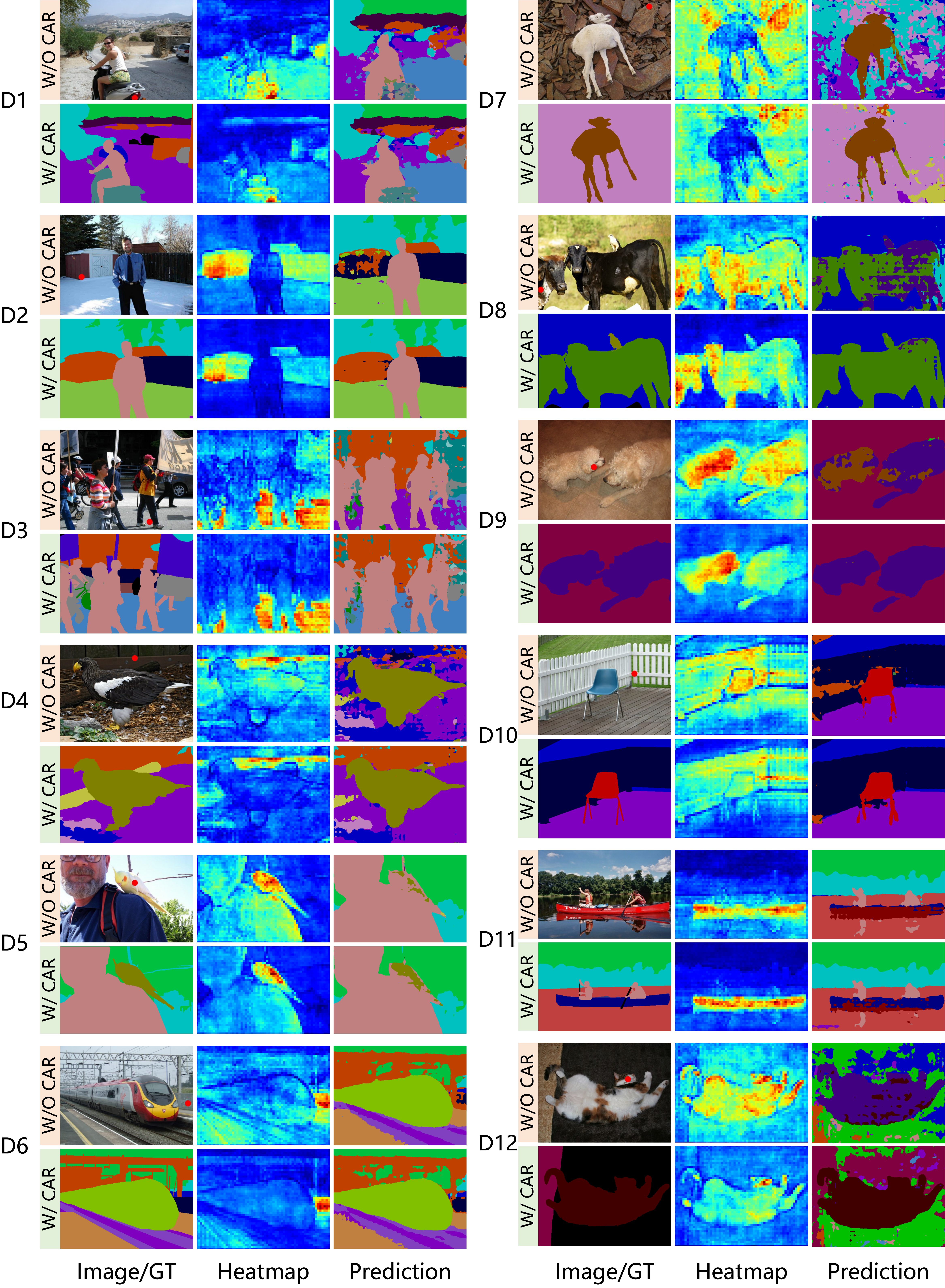}
\caption{Visualization of the feature similarity between a given pixel (marked with a red dot in the image) and all pixels,
as well as the segmentation results of \textbf{ ResNet-50~\cite{cResnet} + DeepLab~\cite{cDeepLabV3}} on Pascal Context test set. 
A hotter color denotes a greater similarity value. 
}
\label{fig:CAR:PixelRelationMap:ResNet50DeepLab}
\end{figure}

We also visualize the pixel-to-pixel relation energy map of ResNet-50~\cite{cResnet} + DeepLabV3~\cite{cDeepLabV3} in Fig.~\ref{fig:CAR:PixelRelationMap:ResNet50DeepLab}. 
These visualizations clearly show that the reduced inter-class dependency helps to correct the classification.

\clearpage

\end{document}